\documentclass[10pt]{article}

\usepackage[margin=1in]{geometry}
\usepackage{booktabs}
\usepackage{graphicx}
\usepackage{amsmath}
\usepackage{amssymb}
\usepackage{hyperref}
\hypersetup{hidelinks,hypertexnames=false}
\usepackage{microtype}
\usepackage{enumitem}
\usepackage{array}
\usepackage{url}
\usepackage{float}
\usepackage{tikz}
\usetikzlibrary{arrows.meta,positioning,calc}
\usepackage{xcolor}

\title{SlideCheck: Guiding Self-Supervised Pretraining of Pathology Foundation Models via Dataset Distributions}
\author{%
Mingyi He$^{1,*}$ \and
Xinyi Guo$^{2,*}$ \and
Xitong Ling$^{3,*}$ \and
Weiming Chen$^{3}$ \and
Jiawen Li$^{3}$ \and
Lianghui Zhu$^{3}$ \and
Minxi Ouyang$^{3}$ \and
Mingxi Fu$^{3}$ \and
Yizhi Wang$^{3}$ \and
Tian Guan$^{3,\dagger}$\\[0.6em]
\small $^{1}$Beijing University of Chemical Technology\\
\small $^{2}$South China Normal University\\
\small $^{3}$Tsinghua University\\[0.3em]
\small $^{*}$Equal contribution. $^{\dagger}$Corresponding author.
}
\date{}

\newcommand{\method}{SlideCheck}
\newcommand{\auc}{\mathrm{AUC}}
\newcommand{\bacc}{\mathrm{BACC}}
\newcommand{\topk}{\mathrm{TopK}}

\begin{document}
\maketitle

\begin{abstract}
Pathology foundation models are pretrained on large streams of WSI-derived patches, while supervision during data construction is often slide-level, sparse, or heterogeneous. This mismatch makes it difficult to understand and control which biological patterns enter the pretraining data. We propose \method{}, a lightweight pretraining data guidance tool built on frozen pathology foundation model patch features. Rather than serving as a standalone patch diagnostic model, \method{} provides explicit abnormality and malignancy scores for organizing, filtering, and auditing pathology pretraining data.

\method{} uses a dual-head MLP to separately model broad abnormal morphology and malignant evidence. A regularized feature-space scorer provides a supervised anchor for patch-level evidence estimation, while score-attention agreement combines patch scores with WSI-level MIL attention to mine high-confidence pseudo labels. The same scores are then used to construct broad-positive ViT pretraining subsets, where a patch is selected if either abnormality or malignancy evidence exceeds a threshold.

Experiments show that \method{}-defined data distributions influence the downstream behavior of self-supervised ViT pretraining, indicating that biological composition is an important controllable factor in pathology foundation model development. Curated subsets can approach full-data performance, suggesting that explicitly scored patch pools may support more efficient and auditable pretraining data construction. These findings position \method{} as a data guidance and auditing layer for transforming large, undifferentiated patch pools into controllable and reusable pretraining datasets.
\end{abstract}

\section{Introduction}

Pathology foundation models increasingly depend on large collections of WSI-derived patches. Recent systems such as UNI~\cite{chen2024towards}, CONCH~\cite{lu2024visual}, Prov-GigaPath~\cite{xu2024whole}, TITAN~\cite{ding2025multimodal}, and Virchow2~\cite{zimmermann2024virchow2} show that scale, diverse histology corpora, slide-level context, and multimodal supervision can produce broadly useful pathology representations. However, the data stream used to build these models is rarely organized at the same granularity as the model input. A slide may be labeled as cancer, but its patches can include malignant epithelium, benign abnormality, normal tissue, stroma, necrosis, inflammation, artifacts, and background. The result is a practical gap between slide-level supervision and patch-level pretraining data, a mismatch also seen in weakly supervised WSI pipelines~\cite{campanella2019clinical,lu2021data}.

This paper addresses that gap from a data-curation perspective. MIL models can classify a bag and highlight contributing instances~\cite{ilse2018attention,lu2021data}, but they do not produce a reusable patch-level interface for organizing unlabeled corpora. A patch classifier produces local disease scores, but as a diagnostic endpoint alone it does not specify how those scores should drive pretraining data construction. Our goal is a scoring layer that bridges the two and makes patch composition explicit before pretraining begins.

We introduce \method{}, a dual-head MLP over frozen pathology foundation model (PFM) features. The abnormality head captures broad non-normal morphology and the cancer head captures malignant evidence; we keep the two heads separate because many pathology patches are abnormal without being cancer. For pretraining curation we then define a \emph{broad-positive} patch as one with either abnormality or cancer evidence above threshold. This definition keeps the curation axis aligned with the ViT subset construction and prevents the sampling variable from collapsing into cancer-only diagnosis.

The proposed workflow is a bridge between three supervision regimes. First, clean patch labels train a regularized feature-space scorer. Second, WSI-level MIL attention is used as an instance prior and intersected with \method{} scores to mine pseudo labels. Third, \method{} scores define controlled self-supervised ViT pretraining subsets for studying sampling ratio, data fraction, and model scale. This design treats sampling ratio as a controlled variable for measuring dataset composition, rather than as a claim of a universally optimal ratio.

\paragraph{Contributions.}
We make four contributions.
\begin{enumerate}[leftmargin=*]
    \item We propose \method{} as an auditable pathology data interface over foundation-model features, not merely as a patch-level cancer detector.
    \item We show that a compact dual-head scorer with feature-space regularization provides a strong supervised anchor for abnormality and cancer scoring.
    \item We use MIL attention as a bag-supervised instance prior and combine it with semantic patch scores to expand patch supervision without rewriting MIL as the final task model.
    \item We use broad-positive \method{} scores to construct ViT pretraining subsets and find that model scale has a larger downstream effect than one-dimensional broad-positive sampling in this study.
\end{enumerate}

\section{Related Work}

\paragraph{Pathology and medical foundation models.}
Computational pathology foundation models have progressed from domain-specific patch encoders and hierarchical WSI transformers toward slide-level and multimodal systems. CTransPath learns a histopathology-specific transformer representation from unlabeled images~\cite{wang2022transformer}, and HIPT scales self-supervised ViTs to gigapixel WSIs through hierarchical pretraining~\cite{chen2022scaling}. More recent systems include UNI as a general-purpose H\&E pathology encoder~\cite{chen2024towards}, CONCH for histology-image and pathology-text alignment~\cite{lu2024visual}, Prov-GigaPath for real-world whole-slide data~\cite{xu2024whole}, TITAN for whole-slide modeling with image-text alignment~\cite{ding2025multimodal}, and Virchow2 for scaled mixed-magnification pretraining~\cite{zimmermann2024virchow2}. Surveys of computational pathology foundation models emphasize that dataset availability, domain variability, adaptation protocols, and evaluation heterogeneity remain major barriers~\cite{li2025survey}. \method{} is complementary to these models: it does not introduce a new large encoder, but supplies a lightweight layer for making patch composition visible and reusable.

\paragraph{Pretraining-data effects and data curation.}
General vision-language and self-supervised work has shown that dataset construction can be a primary research object rather than a hidden preprocessing step. DataComp fixes model training while asking participants to improve the pretraining dataset, Data Filtering Networks show that a model useful for filtering data need not be the same model that performs best on downstream recognition, and DINOv2 emphasizes curated diverse data for robust visual features~\cite{gadre2023datacomp,fang2024data,oquab2023dinov2}. In medical imaging, domain-aligned self-supervised pretraining improves transfer when labels are scarce~\cite{azizi2021big}, and pathology SSL benchmarks show that pretraining choices interact with field of view, stain, and domain composition~\cite{kangbenchmarking}. RETFound demonstrates the value of large-scale retinal self-supervised learning~\cite{zhou2023foundation}, while recent retinal foundation-model studies isolate how pretraining cohorts and demographic attributes affect generalization and fairness~\cite{zhou2026understanding}. Other work explores data-efficient medical foundation models using synthetic disease-conditioned images~\cite{sun2025data}. These studies motivate our central design choice: instead of treating pathology pretraining data as an undifferentiated patch pool, we expose a scoring axis tied to disease morphology.

\paragraph{Weak supervision and MIL in pathology.}
MIL is a natural fit for WSI analysis because labels are often available at bag level while instance labels remain latent. Clinical-scale weakly supervised pathology systems have used reported diagnoses or slide labels to avoid exhaustive pixel-level annotation~\cite{campanella2019clinical}. Attention-based MIL provides permutation-invariant aggregation and instance weights~\cite{ilse2018attention}, and CLAM adapts this idea to data-efficient WSI classification with clustering-constrained instance learning~\cite{lu2021data}. Our use of MIL is intentionally narrower. We use attention as evidence about which patches support a bag-level label, then intersect that evidence with \method{} probabilities. This converts MIL from a standalone classifier into a source of weak instance supervision for patch scoring and data curation.

\paragraph{Self-supervised ViTs.}
Vision Transformers provide scalable image backbones~\cite{dosovitskiy2020image}, while DINO, MAE, and DINOv2 show complementary routes to self-supervised ViT representations~\cite{caron2021emerging,he2022masked,oquab2023dinov2}. In pathology, histology-specific SSL work further shows that objective choice, field of view, stain variation, and domain composition matter~\cite{wang2022transformer,chen2022scaling,kangbenchmarking}. \method{} contributes a simple way to define and audit one clinically meaningful curation axis before pretraining.

\section{Method}

\subsection{Overview: From Weak Slide Labels to Scored Patch Streams}

Let $x_i$ denote a patch feature and $B=\{x_i\}_{i=1}^{n}$ a WSI bag. The central object in \method{} is not a final slide classifier, but a pair of patch scores:
\begin{equation}
    p_i^{\mathrm{abn}} = P(y_i^{\mathrm{abn}}=1 \mid x_i), \qquad
    p_i^{\mathrm{can}} = P(y_i^{\mathrm{can}}=1 \mid x_i),
\end{equation}
where $y^{\mathrm{abn}}$ marks broad abnormal morphology and $y^{\mathrm{can}}$ marks malignant evidence. For pretraining subset construction we use the broad-positive indicator
\begin{equation}
    z_i = \mathbf{1}\left[p_i^{\mathrm{abn}} \geq \tau \;\lor\; p_i^{\mathrm{can}} \geq \tau\right], \quad \tau=0.5.
\end{equation}
A positive pretraining patch under $z_i$ is therefore abnormal-or-cancer; the cancer score $p_i^{\mathrm{can}}$ remains available separately for malignant-evidence analysis, and the ratio induced by $z_i$ should be read as a pathology-evidence ratio rather than a cancer ratio.

\subsection{Feature Representation and Dual-Head Scoring}

For each patch, we use frozen PFM features. The input to the scorer is a fixed-dimensional patch embedding produced by the encoder:
\begin{equation}
  x_i \in \mathbb{R}^{d}.
\end{equation}
Keeping the encoder frozen separates the role of representation learning from the role of data scoring. It also makes the scorer inexpensive to retrain as labels or curation policies change.

\method{} applies LayerNorm, a two-layer MLP, and two binary heads:
\begin{equation}
\begin{split}
 h_i &= f_\theta(x_i), \\
 \ell_i^{\mathrm{abn}} &= w_{\mathrm{abn}}^\top h_i + b_{\mathrm{abn}}, \\
 \ell_i^{\mathrm{can}} &= w_{\mathrm{can}}^\top h_i + b_{\mathrm{can}}.
\end{split}
\end{equation}
The scorer block is
\begin{equation}
\mathrm{LayerNorm}(d) \rightarrow \mathrm{Linear}(d,768) \rightarrow \mathrm{GELU} \rightarrow \mathrm{Dropout} \rightarrow \mathrm{Linear}(768,768) \rightarrow \mathrm{GELU} \rightarrow \mathrm{Dropout}.
\end{equation}
The two heads share representation capacity but preserve a clinically useful hierarchy: malignant patches are abnormal, whereas abnormal patches need not be malignant.

\subsection{Regularizing the Scoring Boundary}

Because the scorer is trained over frozen features rather than raw pixels, regularization is applied directly in feature space. Mixup~\cite{zhang2017mixup} forms convex combinations of features and labels,
\begin{equation}
  \tilde{x} = \lambda x_i + (1-\lambda)x_j, \quad
  \tilde{y} = \lambda y_i + (1-\lambda)y_j,
\end{equation}
with $\lambda$ drawn from a Beta distribution. Feature perturbation independently rescales feature dimensions with small multiplicative noise. These operations smooth the local decision boundary of the MLP without changing the foundation encoder. Empirically, this matters more than increasing the scorer architecture.

\subsection{Attention-Constrained Pseudo-Label Mining}

WSI labels identify bags, not diagnostic patches. To turn bag labels into candidate instance labels, we train a gated attention MIL model~\cite{ilse2018attention,lu2021data} on the same frozen features. For patch $i$ in bag $B$, the MIL model provides an attention weight $a_i$ and the scorer provides $(p_i^{\mathrm{abn}},p_i^{\mathrm{can}})$. High-confidence malignant candidates are selected only when the semantic score and bag attention agree:
\begin{equation}
  \mathcal{P}_{B}^{\mathrm{can}} = \topk_{i\in B}(p_i^{\mathrm{can}}) \cap \topk_{i\in B}(a_i).
\end{equation}
Normal candidates require the opposite pattern: low abnormality, low cancer score, and low attention. Non-malignant abnormal candidates are treated as a separate state, corresponding to high abnormality evidence without high malignant support. This gives three pseudo-label states,
\begin{equation}
    (y^{\mathrm{can}}, y^{\mathrm{abn}}) \in \{(1,1), (0,1), (0,0)\},
\end{equation}
which correspond to malignant abnormal, non-malignant abnormal, and normal tissue. This formulation avoids forcing every suspicious patch into a cancer/non-cancer binary decision and makes explicit why abnormality and cancer need separate heads.

The mined labels are combined with clean supervised patch labels. We evaluate two training routes: finetuning the supervised scorer and training the expanded scorer without supervised initialization. The conceptual role of MIL remains the same in both routes: it supplies an instance-selection prior, while \method{} supplies semantic patch scores.

\subsection{Score-Guided ViT Pretraining Subsets}

The same broad-positive indicator $z_i$ defines self-supervised ViT pretraining subsets. A target ratio $\rho$ samples approximately $\rho$ of the subset from $z_i=1$ patches and $1-\rho$ from $z_i=0$ patches. We use this mechanism to vary three axes: broad-positive ratio, pretraining data fraction, and ViT backbone scale under DINO-style training~\cite{caron2021emerging,kangbenchmarking}. The resulting encoders are evaluated with frozen-feature ROI classification probes. This stage makes pretraining composition an explicit experimental variable without assuming that one positive ratio should dominate across all downstream tasks.

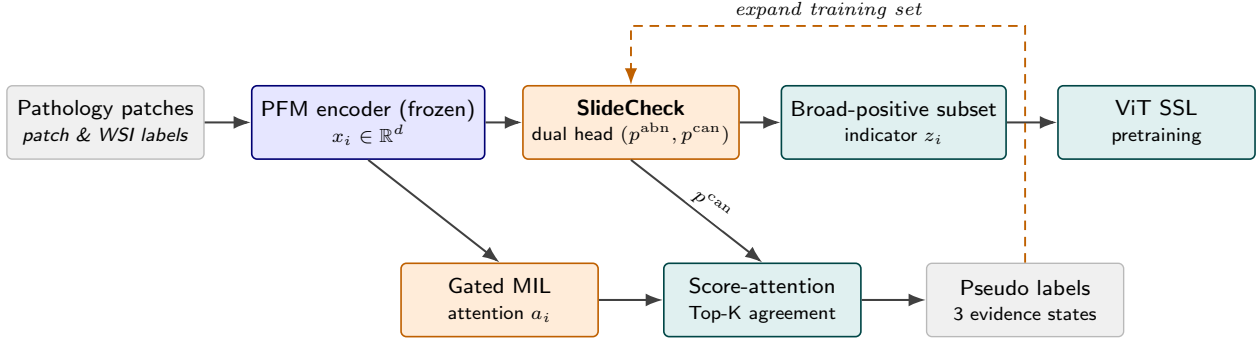
\begin{figure}[t]
\centering
\resizebox{\linewidth}{!}{%
\begin{tikzpicture}[
  font=\sffamily\footnotesize,
  >=Latex,
  databox/.style ={draw=gray!55, fill=gray!12, rounded corners=2pt,
                   minimum width=2.55cm, minimum height=0.95cm, align=center,
                   line width=0.5pt},
  encbox/.style  ={draw=blue!45!black, fill=blue!10, rounded corners=2pt,
                   minimum width=2.55cm, minimum height=0.95cm, align=center,
                   line width=0.6pt},
  modelbox/.style={draw=orange!75!black, fill=orange!15, rounded corners=2pt,
                   minimum width=2.55cm, minimum height=0.95cm, align=center,
                   line width=0.6pt},
  opbox/.style   ={draw=teal!55!black, fill=teal!12, rounded corners=2pt,
                   minimum width=2.55cm, minimum height=0.95cm, align=center,
                   line width=0.6pt},
  flow/.style    ={->, line width=0.75pt, draw=black!75, >=Latex},
  fb/.style      ={->, line width=0.7pt, draw=orange!75!black, >=Latex,
                   dash pattern=on 3.5pt off 2.5pt},
  alab/.style    ={font=\scriptsize, inner sep=1.5pt, fill=white}
]
\node[databox]  (input) at (0,    0) {Pathology patches\\\scriptsize\textit{patch \& WSI labels}};
\node[encbox]   (pfm)   at (3.4,  0) {PFM encoder (frozen)\\\scriptsize $x_i\in\mathbb{R}^{d}$};
\node[modelbox] (slc)   at (6.8,  0) {\textbf{SlideCheck}\\\scriptsize dual head $(p^{\mathrm{abn}}, p^{\mathrm{can}})$};
\node[opbox]    (bp)    at (10.2, 0) {Broad-positive subset\\\scriptsize indicator $z_i$};
\node[opbox]    (vit)   at (13.6, 0) {ViT SSL\\\scriptsize pretraining};

\node[modelbox] (mil)    at (5.1,  -2.3) {Gated MIL\\\scriptsize attention $a_i$};
\node[opbox]    (mining) at (8.5,  -2.3) {Score-attention\\\scriptsize Top-K agreement};
\node[databox]  (pseudo) at (11.9, -2.3) {Pseudo labels\\\scriptsize 3 evidence states};

\draw[flow] (input) -- (pfm);
\draw[flow] (pfm)   -- (slc);
\draw[flow] (slc)   -- (bp);
\draw[flow] (bp)    -- (vit);

\draw[flow] (pfm.south) -- (mil.north);
\draw[flow] (slc.south) -- (mining.north)
  node[alab, pos=0.55, sloped, above] {$p^{\mathrm{can}}$};

\draw[flow] (mil.east)    -- (mining.west);
\draw[flow] (mining.east) -- (pseudo.west);

\draw[fb] (pseudo.north) -- (11.9, 1.25) -- (6.8, 1.25) -- (slc.north);
\node[alab, font=\scriptsize\itshape] at (9.35, 1.45) {expand training set};
\end{tikzpicture}}
\caption{\textbf{\method{} as a patch-scoring interface for pathology data curation.}
Frozen PFM features feed the dual-head \method{} scorer for abnormality and malignancy
and a gated MIL model that consumes WSI bag labels. Score-attention Top-K agreement
mines pseudo labels that expand the \method{} training set (dashed loop).
The broad-positive indicator $z_i$ derived from \method{} scores then constructs
controlled subsets for ViT self-supervised pretraining.}
\label{fig:pipeline}
\end{figure}

\section{Experiments and Analysis}

\subsection{Supervised Scoring on BRACS}

Before any MIL expansion, we evaluate whether the compact scorer alone can provide a reliable patch-level signal. BRACS is a breast H\&E histopathology benchmark with categories that span normal tissue, benign abnormality, atypia, and malignant disease~\cite{brancati2022bracs}, which makes it suitable for evaluating both heads jointly. Table~\ref{tab:slidecheck} compares the base MLP, feature-space regularization, and a larger concatenation variant.

\begin{table}[t]
\centering
\caption{\textbf{BRACS supervised scoring.} Patch-level AUC and BACC for scorer variants; FP denotes feature perturbation. Best values are in \textbf{bold}.}
\label{tab:slidecheck}
\small
\setlength{\tabcolsep}{5pt}
\renewcommand{\arraystretch}{1.08}
\begin{tabular}{@{}lcccc@{}}
\toprule
& \multicolumn{2}{c}{Cancer} & \multicolumn{2}{c}{Abnormal} \\
\cmidrule(lr){2-3}\cmidrule(lr){4-5}
Scorer variant & AUC & BACC & AUC & BACC \\
\midrule
Base MLP                   & 0.9028 & 0.7951 & 0.7394 & 0.6614 \\
\quad + Mixup              & 0.9297 & 0.8568 & 0.9022 & \textbf{0.8343} \\
\quad + FP                 & 0.8845 & 0.8075 & 0.7926 & 0.7011 \\
\quad + Mixup + FP         & \textbf{0.9346} & \textbf{0.8676} & 0.9012 & 0.8212 \\
\addlinespace
Larger concat + Mixup + FP & 0.9172 & 0.8472 & 0.8523 & 0.7682 \\
\bottomrule
\end{tabular}
\end{table}

The regularized compact MLP gives the most reliable BRACS cancer scores, whereas the larger concatenation model does not improve generalization. This supports using \method{} as a practical curation layer: the scorer is inexpensive, transparent, and strong enough to provide an initial abnormality/malignancy signal at corpus scale.

\subsection{Score-Attention Agreement for Expanding Patch Supervision}

We next test whether slide-level supervision can enrich patch supervision. Table~\ref{tab:joint} groups variants by the supervision used to construct pseudo labels. The first row is the supervised patch-label reference; subsequent rows add score-attention agreement under increasingly broad pseudo-label definitions. Evaluation uses BRACS, UNITOPATHO~\cite{barbano2021unitopatho}, and CAMEL cancer detection. The Comp. column summarizes cross-cohort, dual-head performance as
\begin{equation}
\mathrm{Comp.}=\frac{\overline{\auc}_{\mathrm{can}}+\overline{\bacc}_{\mathrm{can}}+\overline{\auc}_{\mathrm{abn}}+\overline{\bacc}_{\mathrm{abn}}}{4},
\end{equation}
where each average is computed across BRACS, UNITOPATHO, and CAMEL.

\begin{table}[H]
\centering
\caption{\textbf{Score-attention expansion.} BRACS, UNITOPATHO, and CAMEL columns report cancer AUC after pseudo-label mining; Comp. is the cross-cohort dual-head composite defined in text. \textsc{ft}/\textsc{scr}: finetune/from scratch; best values are in \textbf{bold}.}
\label{tab:joint}
\small
\setlength{\tabcolsep}{4.5pt}
\renewcommand{\arraystretch}{1.06}
\begin{tabular}{@{}lclcccc@{}}
\toprule
Variant & Init & Pseudo & BRACS & UNITOP. & CAMEL & Comp. \\
\midrule
Supervised & --- & none & \textbf{0.9346} & 0.7977 & 0.8517 & 0.8333 \\
Malig. expansion & \textsc{ft} & malig. & 0.9256 & \textbf{0.8310} & 0.9049 & 0.8531 \\
+ neg. bags & \textsc{ft} & + neg. & 0.9253 & 0.8281 & \textbf{0.9073} & \textbf{0.8591} \\
Dual-head & \textsc{scr} & 3 states & 0.9170 & 0.8257 & 0.9025 & 0.8555 \\
\bottomrule
\end{tabular}
\end{table}

The pattern indicates complementarity: MIL attention supplies bag-level localization, whereas \method{} preserves abnormality and malignancy semantics. Agreement-based expansion improves UNITOPATHO and CAMEL AUC, while clean BRACS patch labels remain the strongest in-domain reference. The dual-head expansion remains competitive without supervised initialization, suggesting that the mined label structure carries reusable signal.

\subsection{Broad-Positive ViT Pretraining Subsets}

\begin{figure}[t]
\centering
\includegraphics[width=0.32\linewidth]{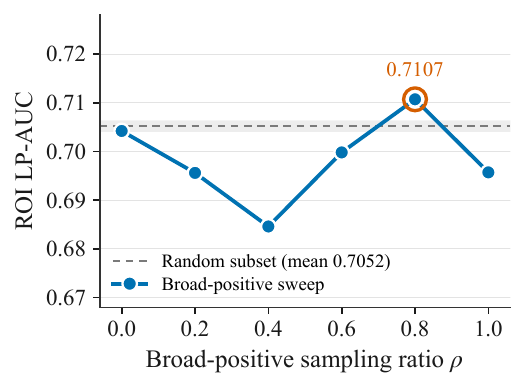}\hfill
\includegraphics[width=0.31\linewidth]{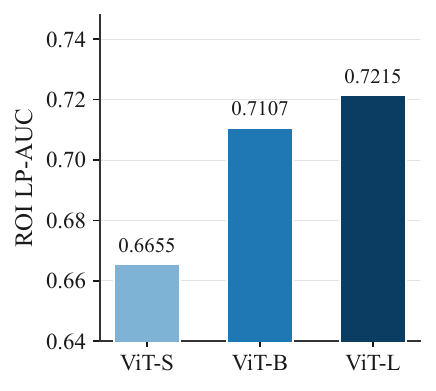}\hfill
\includegraphics[width=0.32\linewidth]{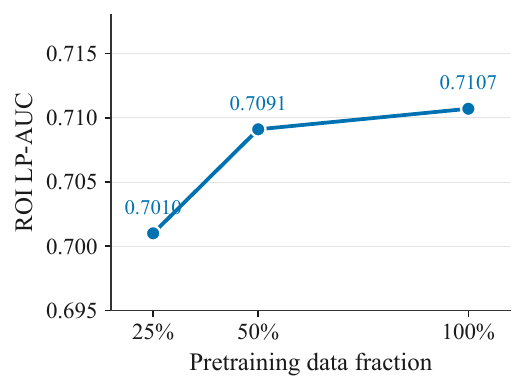}
\caption{\textbf{Downstream behavior under \method{}-guided curation.} Left: broad-positive ratio produces modest variation in ROI LP-AUC. Middle: model scale gives the clearest improvement. Right: smaller curated data fractions approach the full-data ViT-B result.}
\label{fig:vit_curves}
\end{figure}

We then use \method{} scores to form self-supervised ViT pretraining subsets. The sampling variable is the broad-positive ratio $\rho$, where positive means abnormal-or-cancer according to $z_i$, not cancer-only. Table~\ref{tab:vit_summary} summarizes broad-positive ratio, backbone scale, and pretraining data fraction, and Figure~\ref{fig:vit_curves} visualizes the ROI LP-AUC trends. Appendix~\ref{app:fullvit} reports the full run list, including two random-subset baselines.

\begin{table}[H]
\centering
\caption{\textbf{ViT downstream summary.} ROI LP-AUC ranges for the three controlled curation axes.}
\label{tab:vit_summary}
\small
\setlength{\tabcolsep}{5pt}
\renewcommand{\arraystretch}{1.08}
\begin{tabular}{@{}lcccc@{}}
\toprule
Axis & Settings & Min & Best & $\Delta$ \\
\midrule
Broad-positive $\rho$ & 0.0--1.0 & 0.6846 & 0.7107 ($\rho=0.8$) & $+0.0261$ \\
Backbone scale & S/B/L & 0.6655 & 0.7215 (ViT-L) & $+0.0560$ \\
Data fraction & 25/50/100\% & 0.7010 & 0.7107 (100\%) & $+0.0097$ \\
\bottomrule
\end{tabular}
\end{table}

The ViT results support a cautious interpretation. Varying $\rho$ changes LP-AUC by 0.026 across the sweep, whereas moving from ViT-S to ViT-L changes LP-AUC by 0.056 at fixed $\rho$. Thus, \method{} provides a measurable curation axis, but broad-positive ratio alone is not a universal prescription. Future policies can add tissue type, morphology, uncertainty, staining and scanner domain, and slide provenance.

\section{Discussion}

\method{} provides a lightweight layer between pathology foundation encoders and the patch pools used to train or adapt them. Its patch-level abnormality and malignancy scores make pretraining composition explicit: \method{} scores carry morphology, MIL attention links patches back to slide-level supervision, and broad-positive subsets make dataset shifts measurable.

Across the ViT study, backbone scale has a larger LP-AUC effect than broad-positive ratio, and the 25\% curated subset remains within one point of the full-data ViT-B result. These findings support score-guided curation as a mechanism for controlling and auditing pretraining data composition, rather than as evidence for a single optimal sampling ratio.

\paragraph{Limitations.}
The conclusions are restricted to supervised BRACS scoring, score-attention pseudo-label expansion, and DINO-based ViT subset analysis. Seed variability is reported for the random-subset rows but not for each broad-positive ratio, and self-supervised objectives are not compared under a matched design. Abnormal-or-cancer evidence is a one-dimensional curation axis; multi-axis policies along morphology, uncertainty, domain diversity, and slide provenance remain future work.

\paragraph{Reproducibility.}
For each ViT run, the sampling ratio, backbone scale, and data fraction are fixed before downstream evaluation. Appendix~\ref{app:fullvit} lists the runs used to compute Table~\ref{tab:vit_summary}, so the reported ranges and best values trace directly to individual experiments.

\appendix

\section{Full ViT Downstream Table}
\label{app:fullvit}

\begin{table}[H]
\centering
\caption{\textbf{Full ViT results.} Downstream AUC for all DINO-pretrained subsets used in Table~\ref{tab:vit_summary}; \texttt{rand.} denotes uniform sampling.}
\label{tab:fullvit}
\small
\setlength{\tabcolsep}{5pt}
\renewcommand{\arraystretch}{1.06}
\begin{tabular}{@{}lcccrr@{}}
\toprule
Run & Scale & $\rho$ & Frac. & LP-AUC & KNN-AUC \\
\midrule
Rand., seed 42  & ViT-B & \texttt{rand.} & 1.00 & 0.7064 & 0.8577 \\
Rand., seed 123 & ViT-B & \texttt{rand.} & 1.00 & 0.7040 & 0.8595 \\
\addlinespace
$\rho=0.0$ & ViT-B & 0.0 & 1.00 & 0.7042 & 0.8569 \\
$\rho=0.2$ & ViT-B & 0.2 & 1.00 & 0.6956 & 0.8532 \\
$\rho=0.4$ & ViT-B & 0.4 & 1.00 & 0.6846 & 0.8518 \\
$\rho=0.6$ & ViT-B & 0.6 & 1.00 & 0.6998 & 0.8501 \\
$\rho=0.8$ & ViT-B & 0.8 & 1.00 & 0.7107 & 0.8648 \\
$\rho=1.0$ & ViT-B & 1.0 & 1.00 & 0.6957 & 0.8528 \\
\addlinespace
ViT-S, $\rho=0.8$ & ViT-S & 0.8 & 1.00 & 0.6655 & 0.8617 \\
ViT-L, $\rho=0.8$ & ViT-L & 0.8 & 1.00 & 0.7215 & 0.8619 \\
\addlinespace
25\%, $\rho=0.8$ & ViT-B & 0.8 & 0.25 & 0.7010 & 0.8627 \\
50\%, $\rho=0.8$ & ViT-B & 0.8 & 0.50 & 0.7091 & 0.8515 \\
\bottomrule
\end{tabular}
\end{table}

\section{Scope of Claims}

The conclusions are limited to supervised patch scoring, score-attention expansion, and DINO-based ViT subset analysis. SSL-objective comparisons, independent MIL leaderboard rankings, AMD-TB analysis, and t-SNE visualization are outside the matched study design used here.

\bibliographystyle{plain}
\bibliography{references}

\end{document}